# A Novel Topology Optimization Approach using Conditional Deep Learning

**Sharad Rawat and M.-H. Herman Shen[1]**

Department of Mechanical and Aerospace Engineering
The Ohio State University
201 W 19th Avenue
Columbus, OH- 43210

## Abstract

Topology design optimization offers a tremendous opportunity in design and manufacturing freedoms by designing and producing a part from the ground-up without a meaningful initial design as required by conventional shape design optimization approaches. Ideally, with adequate problem statements, to formulate and solve the topology design problem using a standard topology optimization process, such as SIMP (Simplified Isotropic Material with Penalization) is possible. However, in reality, an estimated over thousands of design iterations is often required for just a few design variables, the conventional optimization approach is, in general, impractical or computationally unachievable for real-world applications significantly diluting the development of the topology optimization technology. There is, therefore, a need for a different approach that will be able to optimize the initial design topology effectively and rapidly. In this study, a novel topology optimization approach based on conditional Wasserstein generative adversarial networks (CWGAN) is developed to replicate the conventional topology optimization algorithms in an extremely computationally inexpensive way. CWGAN consists of a generator and a discriminator, both of which are deep convolutional neural networks (CNN). The limited samples of data, quasi-optimal planar structures, needed for training purposes are generated using the conventional topology optimization algorithms. With CWGANs, the topology optimization conditions can be set to a required value before generating samples. CWGAN truncates the global design space by introducing an equality constraint by the designer. The results are validated by generating an optimized planar structure using the conventional algorithms with the same settings. A proof of concept is presented which is known to be the first such illustration of fusion of CWGANs and topology optimization.

## Introduction

Optimization is defined as the process of finding design variables that lead to the best performing function in a set of conditions. Optimization of structures often called Structural Optimization offers a tool, to a designer which utilizes the mathematical algorithms to determine an optimal design having optimal functionality. The optimization function can be a mathematical computable function demonstrating the underlying mechanics of the design problem. However, in most of the real world problems, there is no defined mathematical function which relates the design variable to the concerned response. In cases where either mathematical model is not available or requires heavy computations, an approximate meta model, often called a surrogate model is evaluated from a set of data points. These surrogate models exhibit the mechanics of the problem. Established optimization techniques, gradient as well as non-gradient techniques [1-3] can be used to find the optimal design variables based on these surrogate models.

One such optimization technique is topology optimization which is an optimal material layout problem [4-6]. Topology optimization is different from size and shape optimization because size and shape optimization techniques are unable to generate a new topology of a structure. Since topology of a structure heavily influences the functionality of the structure, topology optimization is a valuable preprocessing tool for designers. The design iterations begin with a uniform density throughout the design domain followed by a finite element analysis. The densities are altered to ensure a fixed volume constrained given for the design and to minimize the objective the function. Since densities are allowed to either be 0 (no material) or 1 (with the material), an approach called Solid Isotropic Material with Penalization (SIMP) [4], also called the power law approach, is utilized where the intermediate densities are penalized to limit them to 0 or 1 value. There have been many other ways to perform topology optimization namely; Homogenization based optimization (HBO) where the density values of 0/1 are homogenized and non-gradient based algorithms which include evolutionary structural optimization (ESO), and bi-directional ESO (BESO) [7] have been applied in real-world applications.

Topology optimization has been extensively applied to diverse optimization problems such as compliance minimizations and thermal engineering problems in a variety of industries. However, this approach also brings an inherent limitation to the process. Topology optimization demands a high-efficiency computational infrastructure. For large design problems, this approach has exorbitant computational costs which lead to time delays in an iterative design process. There have been research efforts to reduce the computational requirements without compromising on the quality of the structures. Kim et al. [8] introduce an approach to reduce the computational time by reducing the adaptively reduce the design variables which converge quickly during the design process. This led to a faster convergence. Liu et al. [9] propose to reduce the computational time by reducing the dimension of the design variable using an extremely popular unsupervised machine learning algorithm called K-means algorithm. Wu et al.[10] reduced the computational time by proposing a system equipped with high-performance GPU solver which is able to handle large 3D design problems efficiently.

Recently, with the advent of high-performance computing hardware and achievements in deep learning, artificial neural networks are being applied in wide variety of problems such as in self-driving cars, natural language processing, computer visions, healthcare, and financial sector. Artificial neural networks are capable to develop a non-linear function between the variables and the objective. This is exploited by researchers to avoid running a computationally heavy simulation in engineering [11-14]. In the field of design, machine learning algorithms have been applied for design as well as manufacturing [15]. In the area of topology optimization, deep learning was applied by Sosnovik and Oseledets [16] where a deep convolutional neural network was employed to map the intermediate topology optimization result to the converged optimal topology. This speeds the process of convergence of the optimization. Since this approach converts the topology optimization problem from a design problem to an image segmentation problem, this technique lacks an integration of the design optimization settings into the problem solution. Convolutional neural networks are used for classification and regression techniques only and are unable to produce new unseen structures. However, recently [17], generative adversarial networks (GAN), a class of generative models have been invented which have the ability to generate new images or optimal structures if applied to design problems. A special class of GANs is conditional GAN (CGAN) [18] where the optimization design domain is segmented into the different spaces depending on the constraints. CGAN provide an ability to get the desired features in the generated samples which was a limitation of GANs. Yu et al [19] [20] used a generative model called Variational Auto encoder (VAE) to generate the sub-optimal structures and utilized GANs to improve the quality of the structures by increasing the resolution of the generated structure (using a Super Resolution GAN [21]). However, GAN's ability to generate high quality new optimal structures was not exploited. A new WGAN-CAE architecture was proposed by Li et al. [22] where GANs were employed to generate new samples from a latent space of a deep autoencoder. This work also employed a convolutional neural network to classify the generated samples into their design optimization parameters. Rawat et al [23] implemented a generative network to generate quasi-optimal planar structures with a small dataset of 3024 samples. However, the trained GAN was unable to generate the planar structure of desired optimization conditions. Their article utilized a deep convolutional network to determine the optimization parameters. The current article is an extension of their work. In this article, a conditional WGAN is implemented to solve the topology optimization problem.

Conditional WGANs (CWGAN) provide an equality constraint to the optimization problem. In this study, the constraint is defined as the desired volume density in the planar structure. CWGAN generate a high-quality sample but also bring a set of limitations as explained in this article. Section 2 discusses the design problem and the existing method (SIMP) to solve the problem. Section 3 elaborates on GAN and CWGAN along with the architecture of CWGAN used in this study. Section 3 also explains the method for generation of the dataset. Finally, in section 4 results and inferences from the outcomes of this study are shown. The discussions in section 4 provide explain the limitations of the current framework. Comments, future work and concluding remarks are given in the last section. This is one of the first implementations of CWGANs for generation of quasi-optimal structure from noise results from which can lead to a significant time and cost reduction of the design process.

## Problem Statement

Topology optimization is a systematic method to optimally distribute material in a design space under given constraints and conditions to attain the optimal characteristics of the structure [24]. The defined conditions include the loading conditions, boundary conditions along with the volume density required in the desired structure.

Rozvany et al [25] proposed the famous SIMP algorithm to achieve practical designs for an optimization problem. Figure 1 shows the flow chart for the SIMP algorithm. In SIMP, a penalization factor *penal* is introduced which penalizes the fractional densities and hence limiting them to 0 or 1. SIMP is popular due to its ease in implementation and conceptual simplicity.

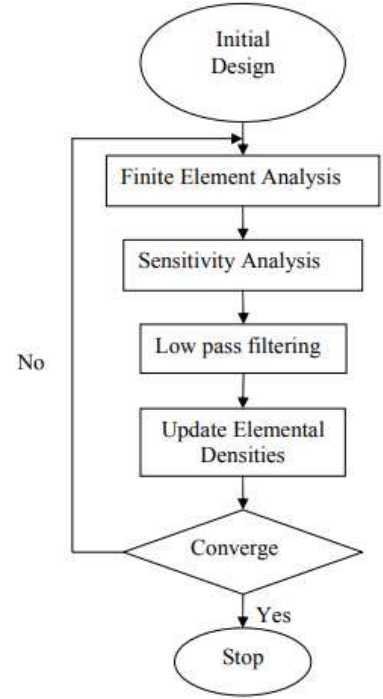

Figure1. Flowchart of Topology Optimization using SIMP [26].

Mathematically, the topology optimization implementation using SIMP can be defined as:

$$\min_{x} : \quad c(x) = U^T K U = \sum_{e=1}^{N}(x_e)^p\, u_e k_o u_e$$

$$subjected\ to: \quad \frac{V(x)}{V_0} = f,$$

$$KU = F,$$

$$0 < x_{min} \leq x \leq 1$$

Where c(x) is the objective function, U and F are the global displacement and force vectors respectively. K is the global stiffness matrix, $u_e$ and $k_o$ is the element displacement vector and stiffness matrix, respectively, x is the vector of design variables, $x_{\min}$ is a vector of minimum relative densities. These should be non-zero to avoid singularities. N (= nelx×nely) is the number of elements used to discretize the design domain, p is the penalization power, V (x) and $V_o$ is the material volume and design domain volume, respectively and f (volfrac) is the prescribed volume fraction [27] [28]. This existing method is used to generate a dataset of 3024 samples as discussed in Section 3. Although topology optimization as a tool provides an initial optimal design for an iterative design process, it has a huge computational cost which increases exponentially with mesh increase in dimensions or increase in the mesh discretization. Therefore, the authors extend their study of integrating deep learning to topology optimization [23], to conditional WGANs which are explained in the next section.

## Deep learning for Topology Optimization

Generative adversarial Network (GAN) was a recent breakthrough by Goodfellow [17] which have the ability to generate new unseen data. GANs consist of 2 networks, a generator G (z) and a discriminator D(x). Generator and discriminator networks are adversaries of each other during the training process. The discriminator is trained to successfully distinguish between the true and fake images. Whereas, the generator is trained to generate a realistic image from noise such that the discriminator is unable to distinguish between the real and fake images. Mathematically, the objective function is described as:

$$\min_G \max_D V(D,G) = E_{x \sim P_{data}}[\log(D(x)] + E_{z \sim P_z}[\log(1 - D(G(z)))]$$

Where x is the image from training dataset $P_{data}$ and z is the noise vector sampled from $P_z$.

During the training of Generative adversarial networks search for Nash Equilibrium of a non-convex game with various parameters is conducted. This leads to a highly unstable learning [29]. To improve learning a Wasserstein Generative Adversarial Network (WGAN) [30] is used instead of a conventional GAN there is a minimax game between a generator and a discriminator. WGAN is different from conventional GANs in its objective function. WGAN utilizes earth mover distance (EM) instead of Jenson-Shannon Divergence in GANs. Deep convolutional networks have also been integrated with GANs to improve training [29]. Deep convolutional neural networks have been explained in detail later in this section. The generator is trying to fool the discriminator and the discriminator attempts to identify fake images. In the training of the proposed model, the discriminator is first trained through the dataset of true images from SIMP model followed by generated images from the generator. Therefore, the generator improves its ability generate true-like fake images from the noise and the discriminator improves its ability to detect the image as real or fake until the discriminator is unable to detect the difference between a real and fake image.

Conditional GAN is a class of GANs where a constraint is added to the optimization problem. In GAN, there is no control over modes of the data to be generated. The conditional GAN, therefore, adds a label as an additional parameter to the generator and this label can be altered once the training is complete to generate a diversity of characteristics in a structure. Labels are also added to the discriminator input to distinguish real images better. This divides the design space in accordance with different constraints. For example, in this article the labels of volume fraction are used as constraints, therefore, in each epoch constraints are set as the corresponding volume fraction for each image in the batch. This constraint focuses the GANs to learn the conditional distribution at each constraint. Hence, this leads to a better approximation of the complex true distribution. Therefore, the authors expect better quality structures from CWGANs. The objective function for Conditional GANs is different from GAN's objective function. Mathematically the objective function of CGAN is:

$$\min_G \max_D V(D,G) = E_{x \sim P_{data}}[\log(D(x|y)] + E_{z \sim P_z}[\log(1 - D(G(z|y)))]$$

For CWGAN:

$$\min_G \max_D V(D,G) = E_{x \sim P_{data}}[D(x|y)] - E_{z \sim P_z}[D(G(z|y))]$$

Where y is the labels corresponding to the data supplied to the problem. Other variables have the same meaning as described previously.

Figure 2 represents the basic flow of data in deep conventional DC Conditional GAN. In Conditional GANs, an image and the corresponding label is required as inputs to the discriminator as can be seen in figure.2. Inputs to the generator are derived from noise. The output of the generator is an image and its corresponding label which is supplied to the discriminator. Discriminator trained from the real images and corresponding labels classifies the generated image into real or fake. This response is propagated back to the generator to adjust the weights of the neural network of the generator. For future references in this paper, all the mentions of GANs assume they are an integration of convolutional neural networks and generative adversarial network. The architecture CWGAN is discussed in detail later in the section.

## Convolutional Neural Network

Convolutional neural networks (CNN) are deep artificial neural networks that are primarily used for image classification, cluster them into classes or perform a regression mapping data. These networks understand and remember the correlation between pixels or voxels in an image/3d structure and use this knowledge to classify the new data. LeNet, proposed by Lecun [31], is one of the first breakthrough CNN architectures. CNNs have been proved to be efficient while working with images. Therefore, it is logical to include CNN in GANs when the dataset contains images.

## Word Embedding

Word embedding is a set of techniques where the data in the design space is mapped to another vector of another dimension. Word embedding is an extremely powerful tool to relate the data having different dimensions. These techniques are primarily common in natural language processing where word embedding encodes the similarity such that embedding for similar words is nearby to each other in the new vector space [32]. In the present study, in order to map labels to the planar structures, word embedding is employed. Before entering the CWGANs architecture, the label is mapped to dimensions of the planar structure. Once of the same dimension, the planar structure image and the modified labels are multiplied, hence embedding the label into the planar structure image. The product of the multiplication has used the input to the CWGANs.

## The architecture of the Generative adversarial network

As mentioned in the previous section (Introduction), GANs used in this article have been modified over the vanilla GAN architecture. Firstly, a deep convolutional GAN (DCGAN) has proved to be producing realistic images. Therefore, the generator and discriminator have a deep convolutional neural network architecture. In an attempt to stabilize the training, Wasserstein GAN (WGAN) is used where the objective function is modified and the weights are clipped to enforce the Lipchitz constraint.

regularize the objective function and are added to reduce the overfitting during training. Batch Normalization [35] is a technique to normalize the samples between the layers while training. This helps in better learning by avoiding the problem of vanishing gradient. The numbers on the left and right side of both the images indicate the size of the tensor and the depth of that tensor at that layer respectively. For the discriminator, the input size is a tensor of 120x120. The output of the discriminator is a 1x1 tensor between -1 (real images) and 1 (fake images). For the generator, the input is a noise of latent dimension of size 120x1. After repeated convolutions layers and batch normalization layers, the output from the generator is a tensor of size 120x120. The labels are integrated into both the architectures by using word embeddings. The latent information of the data is mapped to the corresponding volume fraction. The discriminator and the generator are alternatively trained. RMSProp is used as the optimizer with a learning rate of 0.00005. An activation function of Leaky ReLU with an alpha of 0.2 is used. To avoid unstable training, an attempt has been made to clip the weights of the discriminator and to smoothen the labels on both sides i.e. the labels were changed to (-0.9, 0.9) instead of (-1, 0) [29]. The architecture was created using Tensorflow backend Keras library in python 3.5.

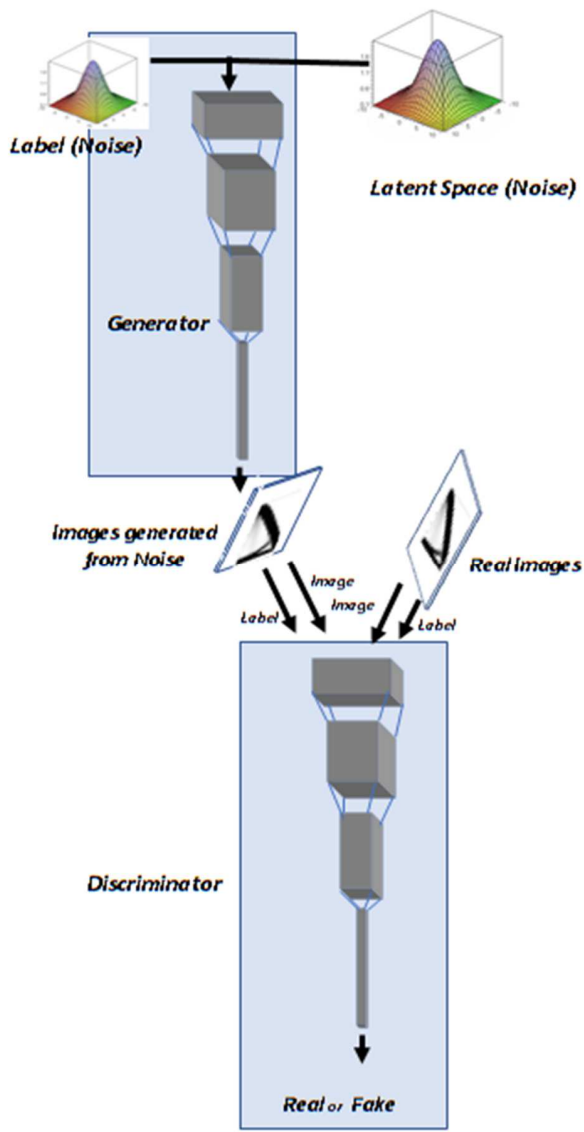

Fig.2 Basic architecture and training process of a DC Conditional GAN.

The architecture for CWGANs used for this study is shown in Fig. 3 and Fig. 4. This architecture consists of convolutional neural networks in generators as well as the discriminator. The generator is deeper than the discriminator. No pooling layer was either in discriminator or the generator [33]. Dropout layers [34] are a way to

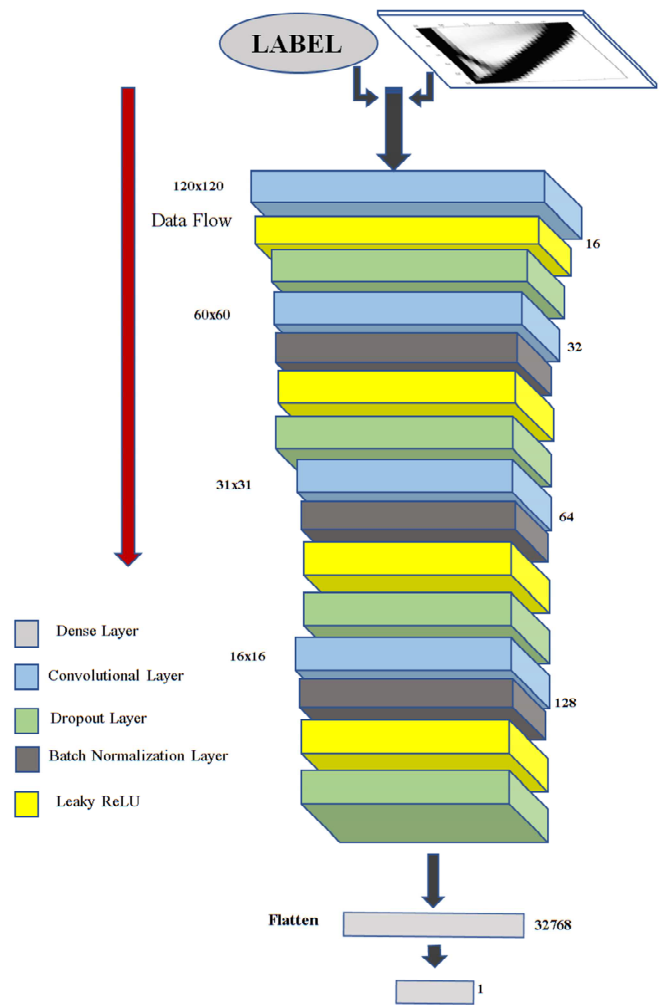

Fig.3 Architecture of discriminator of the CWGAN used.

## Dataset

A conventional SIMP model topology optimization algorithm is used to generate dataset in MATLAB. A dataset of 3024 samples was generated using volume fraction (vol_frac), penalty (penal) and radius of the smoothening filter ($r_{min}$) as design variables. Following are bounds of each variable:

1) Volume Fraction (vol_frac) – 0.3-0.7 (increment of 0.05)
2) Penalty (penal) – 2-4 (increment of 0.1)
3) Radius of filter (rmin) – 1.5-3 (increment of 0.1)

There can be more variables to this problem but since this article is a proof of concept, only 3 scalar variables are chosen. Each dataset was created at a resolution of 120 X 120 pixels with same boundary condition of a cantilever beam loaded at the mid-point of hanging end. The dataset generated is used to train the CWGAN.

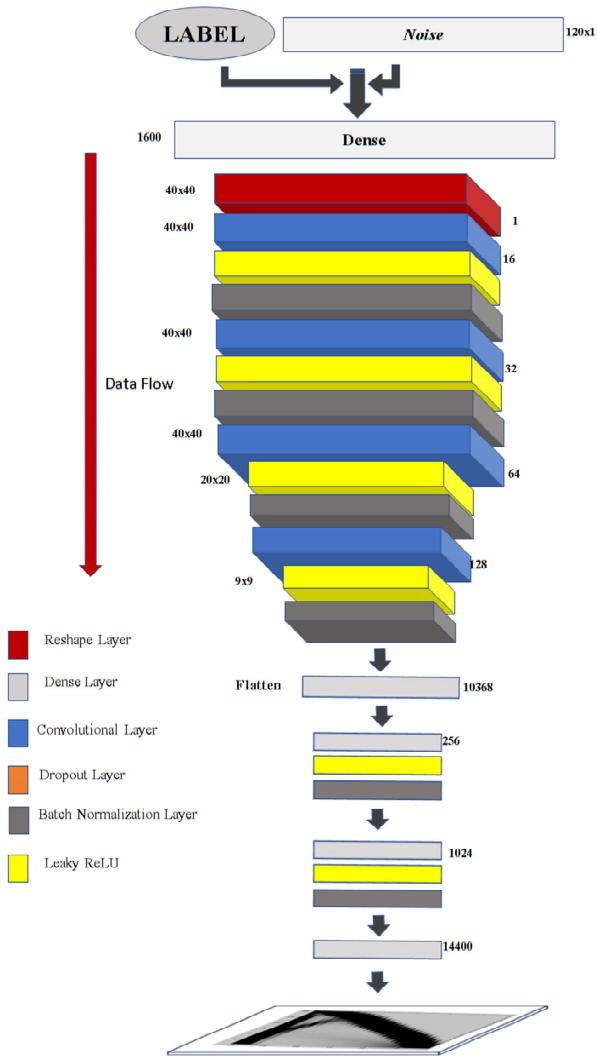

Fig. 4 Architecture of the generator used in CWGAN

## Results and Discussions

In this study, conditional generative adversarial networks generated quasi-optimal design structures for a given boundary condition and optimization settings. In contrast to the GANs, in CWGANs, a planar structure with desired characteristics can be generated from the model. In this study, only the volume fraction is chosen as the conditioning label. Therefore, the model should generate the planar structure corresponding to the defined volume fraction by the user. In this section, the structures obtained from CWGAN are presented along with the inferences from the results.

A conventional topological optimization algorithm is used to generate a small number of datasets as displayed by Figure 5. Samples taken from the datasets are supplied to the discriminator network in CWGAN for training. Figure 6 shows the samples generated from CWGAN after training. An investigation is done corresponding to the volume fraction of 0.4. It can be seen that CWGANs have been able to replicate the distribution of dataset (Pdata), at least for 0.4 volume fraction.

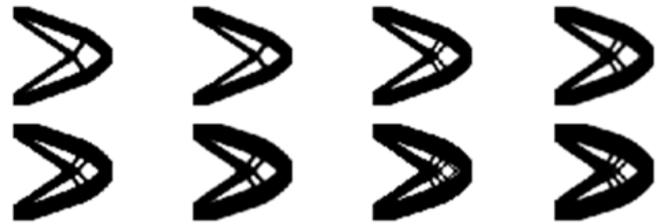

Fig.5 Optimal structures from the conventional algorithm

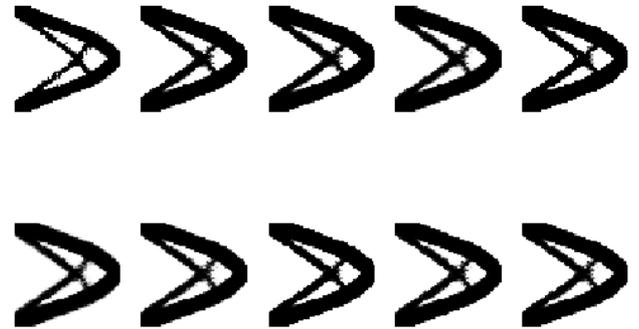

Fig. 6 Unprocessed quasi-optimal structures generated from CWGAN

## Post-processing

To further improve the quality of the planar structures, a post-processing was necessary. A threshold filter where all the pixels above a value of 0.5 are rounded to 1 and pixel values below 0.5 are rounded to 0 is applied. This filter improves the quality of the structure but at the same time sharpens the resulting structure. To further smoothen the planar structure, a bivariate Gaussian filter with a kernel of 5, in both dimensions, is applied to the structure.

A further investigation is conducted to check the robustness of the model at different volume fractions. However, the model gives poor results at volume fractions other than 0.4. Table.1 presents the comparative results of CWGANs against the planar structure obtained from the conventional algorithm for a variety of volume fractions.

Results from table.1 demonstrate that irrespective of the volume fraction, the generated structure's volume fraction was close to 0.4. This is an issue of the existing model. This lack of diversity in the generated output is called Mode collapse. Mode Collapse is a condition where the generator collapses to a setting where it produces similar outputs. Although with CWGAN, the objective is earthmover's distance and a non-normal distribution of the data is assumed by the model, the generator of CWGAN yet collapses. This remains unknown to our knowledge. A potential explanation to mode collapse is overtraining. However, CWGANs exhibit 2 kinds of divergence:
- a) Not following the condition
- b) Not converging to a structure

Table 1. Comparative results from CWGAN and the corresponding conventional algorithm

| CWGAN result | SIMP result |
|---|---|
| vol_frac - 0.3 | |
| Time: 0.350 s | Time: 83.914 s |
| 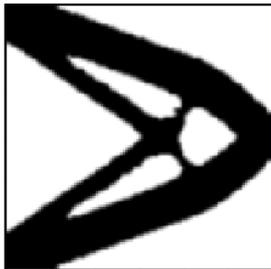 | 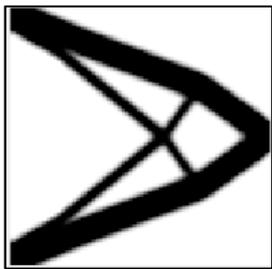 |
| vol_frac – 0.4. | |
| Time: 0.34070 s | Time: 72.318 s |
| 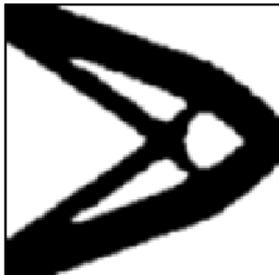 | 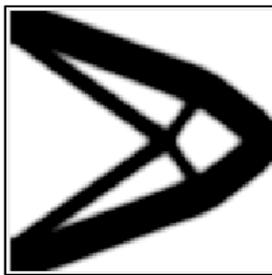 |
| vol_frac – 0.5 | |
| Time: 0.339 s | Time: 71.461 s |
| 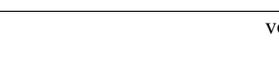 | 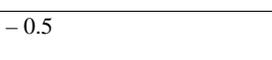 |
| vol_frac – 0.6 | |
| Time: 0.353 s | Time: 114.635 s |
| 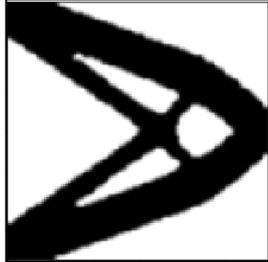 | 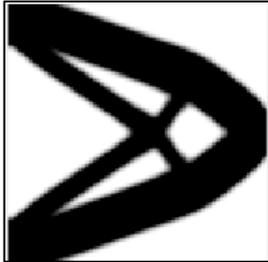 |
| vol_frac – 0.7 | |
| Time: 0.347 s | Time: 83.613 s |
| 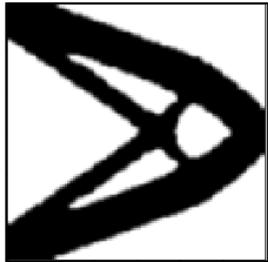 | 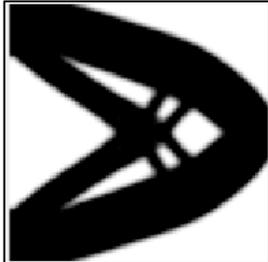 |

Not training the model for long results in either of these divergences. Therefore, there's no guideline about the required epochs to stop for CWGANs. The guideline of stopping at where the output probability of discriminator reaches 0.5 doesn't work when handling CWGAN models. This is still an ongoing research and authors would extend their work addressing these limitations.

Table1 also presents the computation costs of producing each of the structure. In terms of the computational efficiency, results show remarkable reductions in the computational time of an order of $10^2$. For example, the time taken to compute a structure from a trained CW GAN is 0.350 seconds against 83.914 seconds taken by the traditional SIMP approach. Therefore, CWGAN provides an efficient tool for designers to iterate between the initial designs quickly without losing the optimality.

## Conclusions

Conditional GANs are an important class of GANS. These networks provide control for getting the quasi-optimal structures with desired characteristics which were lacking in GANs. In this study, CWGANs generate quasi-optimal structures for a pre-defined set of constraints and conditions. On comparing the quality of structures against the structures generated from SIMP, CWGANs show a capability of generating a quasi-optimal topology once trained with a very small dataset of 3024 samples. The CWGANs give a good result for a single constraint of volume fraction equal to 0.4. However, mode collapse was observed for other constraints, which is responsible for the lack of diversity in the generated output structures. Further investigation is recommended in this area. GANs also lack a proper evaluation metric which makes it difficult to evaluate the results generated by the GANs. The authors have noted these limitations of GANs and hope to extend this work by researching into these problems. The conditional GAN is an equality constraint optimization problem and tools like Lagrange multiplier could be included to handle the equality constraint. Therefore, reconstruction of the objective function is another potential research track.

This work combines the capabilities of deep learning with mechanical design processes. This work will transform the design optimization technology for topology optimization. This method aids designers in reaching an initial design for the design process in a very computationally inexpensive way.

## Contact Information

MH Herman Shen, shen.1@osu.edu, +1(614)- 292-2280

## Acknowledgment

The authors thank the Ohio Supercomputer Center for providing High-Performance Computing resources and expertise to the authors. OSC is a member of the Ohio Technology Consortium, a division of the Ohio Department of Higher Education.

## Definitions/Abbreviations

| | |
|---|---|
| **GAN** | Generative Adversarial Networks |
| **CGAN** | Conditional Generative Adversarial Networks |
| **WGAN** | Wasserstein Generative Adversarial Networks |
| **CNN** | Convolutional Neural Networks |
| **DCGAN** | Deep Convolutional Generative Adversarial Networks |
| **SIMP** | Simplified Isotropic Material with Penalization |
| **VAE** | Variational Autoencoders |
| **CWGAN** | Conditional Wasserstein Generative Adversarial Networks |